\pdfoutput=1
\documentclass{whitepaper}

\usepackage{amsmath,amsfonts,amssymb,bm}
\usepackage{xspace}
\usepackage{wrapfig}
\usepackage{makecell}
\usepackage{colortbl}
\usepackage{float}          %

\crefname{figure}{Fig.}{Figs.}
\crefname{table}{Tab.}{Tabs.}
\crefname{equation}{Eq.}{Eqs.}
\crefname{section}{Sec.}{Secs.}

\newcommand{\eq}[1]{Eq.~(\ref{#1})}

\newcommand{\eg}{\textit{e.g.}\xspace}

\definecolor{colorfirst}{rgb}{0.7, 1, 0.7}   %
\definecolor{colorsecond}{rgb}{1, 1, 0.7}    %
\definecolor{colorthird}{rgb}{1, 0.7, 0.7}   %

\newcommand{\cellfirst}{\cellcolor{colorfirst}}
\newcommand{\cellsecond}{\cellcolor{colorsecond}}
\newcommand{\cellthird}{\cellcolor{colorthird}}

\newcommand{\cmark}{\textcolor{green!55!black}{\checkmark}}
\newcommand{\xmark}{\textcolor{red!65!black}{\ensuremath{\times}}}

\newcommand{\gridheaders}[4]{%
  {\sffamily\small
   \makebox[0.25\linewidth][c]{#1}%
   \makebox[0.25\linewidth][c]{#2}%
   \makebox[0.25\linewidth][c]{#3}%
   \makebox[0.25\linewidth][c]{#4}\par}%
  \vspace{1pt}%
}

\begin{document}
\nocite{*}

\thispagestyle{plain}
\begin{center}
  {\centering\includegraphics[height=11mm]{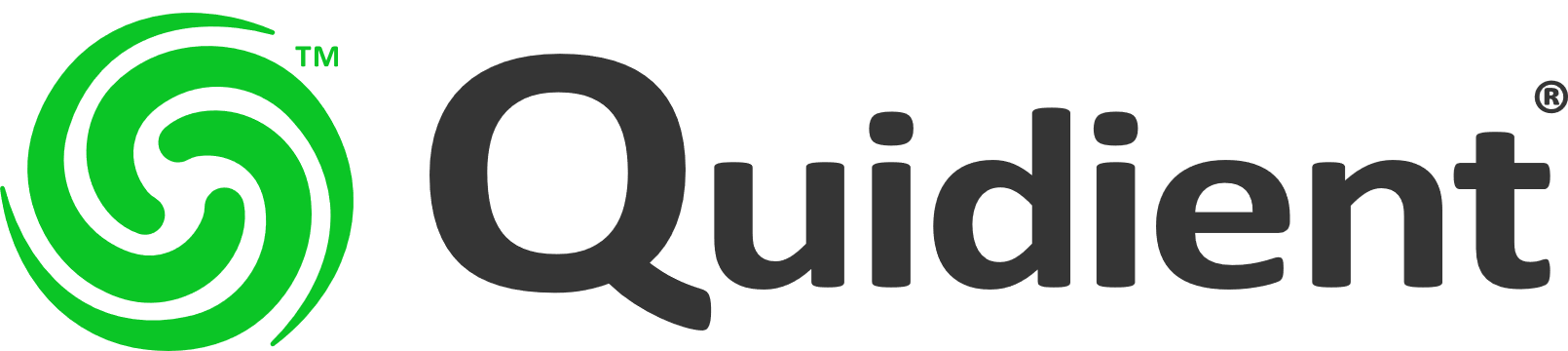}\par}
  \vskip 3mm
  {\sffamily\fontsize{13}{16}\selectfont Quidient Tech Note\par}
  \vskip 6mm
  {\sffamily\bfseries\fontsize{23}{27}\selectfont Plenoptic Condensation\par}
  \vskip 2mm
  {\sffamily\fontsize{15}{19}\selectfont A Novel Approach to Generalized Scene Reconstruction\par}
  \vskip 6mm
  {\sffamily\fontsize{12.5}{16}\selectfont
    Brevin Tilmon, Alex DeJournett, John Leffingwell, Scott Ackerson\par}
  \vskip 2mm
  {\sffamily\fontsize{11}{14}\selectfont Quidient, Columbia, MD\par}
\end{center}

\vskip 4mm

\begin{center}
  {\sffamily\bfseries\color{accent}\fontsize{12}{14}\selectfont Abstract\par}
\end{center}
\vskip 1mm
{\noindent
We present a novel Generalized Scene Reconstruction (GSR) approach called Plenoptic
Condensation (PCon). PCon uses a multi-stage reconstruction pipeline, initially
converting images into ``soupy'' scene elements with low (representational) power,
then adaptively \emph{condensing} the ``soup'' into ``structured'' elements of higher
power capable of efficiently representing, for example, sharp edges and smooth
reflective surfaces. PCon scene models called Reality Models\texttrademark{} (Relms)
enable spatially varying representational power, which is essential for high-fidelity
rendering, measurement, and scene understanding. We showcase several in-the-wild PCon
reconstructions captured with consumer phone cameras and drones. In one case called ``Damaged
Fiat,'' PCon is benchmarked against two state-of-the-art (SOTA) GSR methods: NeRO and
RT-Splatting. Referring to Figure 1 below, PCon reconstructs the car hood
more than twice as accurately as the SOTA methods. But more importantly,
the local damage profile error for PCon is 35\,$\mu$m (0.035 mm), whereas the two other
SOTA methods are essentially unable to measure the damage at all. Our project website is
available at \url{https://quidient.github.io/pcon-2026}.\par}

\vspace{2mm}
\noindent\begin{minipage}{\linewidth}
  {\centering
    \begin{minipage}{0.62\linewidth}
      \gridheaders{Ground Truth}{PCon}{NeRO}{RT-Splatting}
      \includegraphics[width=\linewidth]{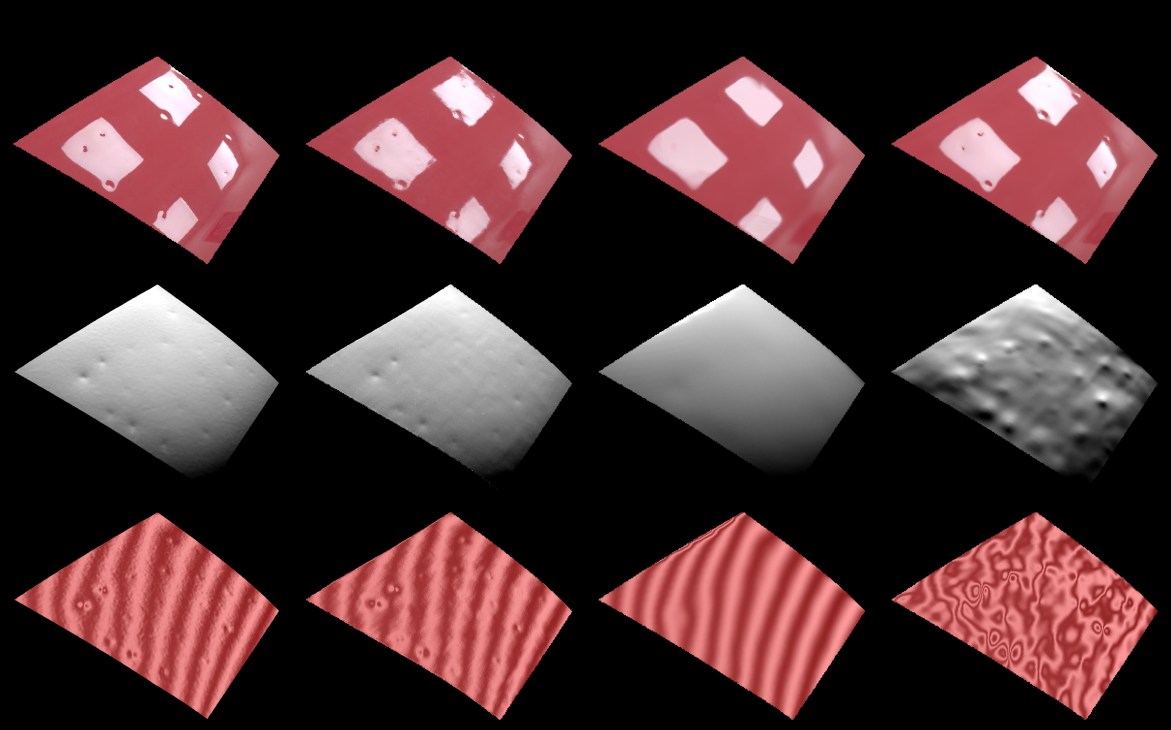}
    \end{minipage}\par}
  \vspace{1mm}
  \captionof{figure}{We compare PCon to two state-of-the-art generalized scene
    reconstruction (GSR) methods: NeRO and RT-Splatting. The top row shows real (GT) or
    rendered (GSR) images. The middle row shows the virtually illuminated mesh surface
    normals. The bottom row shows the mesh illuminated by a virtual stripe projector
    with a manual BSDF that makes the dents easily perceived by humans.}
  \label{fig:teaser}
\end{minipage}

\clearpage
\renewcommand{\cfttoctitlefont}{\sffamily\sectionfont\color{accent}}%
\renewcommand{\cftsecfont}{\sffamily\color{accent}}%
\renewcommand{\cftsecpagefont}{\sffamily\color{accent}}%
\renewcommand{\cftsubsecfont}{\color{black}}%
\renewcommand{\cftsubsecpagefont}{\color{black}}%
\begingroup
  \hypersetup{hidelinks}%
  \tableofcontents
\endgroup
\clearpage

\section{About this Tech Note}
\label{sec:about}

\subsection{Terminology}
\label{sec:terminology}

\textbf{Generalized Scene Reconstruction.} Generalized Scene
Reconstruction\footnote{``Generalized Scene Reconstruction'' is a descriptive term of
art which is not subject to trademark.} (GSR) is an important
branch of Spatial AI focused on creating immersive scene models from images, where the
models enable light and matter to be at least partially disentangled\footnote{Loosely
speaking, GSR methods use a minimum of five coordinates (x, y, z, i, j) to represent
scene elements in contrast to older 3D methods that use a minimum of three (x, y, z).
The extra coordinates are used to represent light flowing in a scene. One can say when
contrasting GSR to earlier 3D methods that ``5D is the new 3D.''}. NeRFs, Splats, and
Relms are pioneering GSR methods.

\subsection{Contributions}
\label{sec:contributions}

We describe a novel approach to Generalized Scene Reconstruction (GSR) called Plenoptic
Condensation (PCon). Our contributions are:
\begin{itemize}
  \item \textbf{High-fidelity immersive models.} PCon jointly condenses a scene's
        matter field and light field into a Reality Model (``Relm'') with high fidelity
        (accuracy, resolution, and representational power). PCon reconstructs
        appearance, geometry, PBR materials (\eg, for relighting), and light field
        information jointly, rather than requiring separate systems (Sec.~4).
  \item \textbf{Relightable PBR mesh export.} PCon exports relightable,
        pipeline-ready PBR meshes as a low-loss export of the native Relm, not a lossy,
        entangled (``baked-in'') export.
  \item \textbf{Accuracy on demand.} A mixed-primitive representation provides
        high-accuracy geometry where it is needed (say 1 part in 1,000 to 1 part in
        10,000) (Sec.~4.6).
  \item \textbf{5D rest of scene.} A single scene model recovers near-field and
        far-field scene elements in the regions beyond the camera workspace (Sec.~4.8).
  \item \textbf{Benchmark SOTA GSR approaches.} Using a smartphone to capture a Fiat
        500 in an indoor space, we benchmark PCon performance against NeRO and
        RT-Splatting using metrology-grade ground truth and a common protocol (Sec.~5).
\end{itemize}
The first part of this Tech Note (Secs.~1 and 2) frames why GSR matters and where it is
used. The remainder (Secs.~3--6) details how PCon condenses Relms and quantifies
its performance.

We demonstrate PCon operation on a deliberately challenging real-world scene: a damaged
automobile body panel\footnote{Hood of a red 2012 Fiat 500 Pop with small dents.} under
uncontrolled indoor lighting. The panel contains several shallow dents whose
reconstruction demands accurate disentanglement of matter, light, and their interaction
with each other (BSDF). We captured image data of the panel using a consumer phone camera
and ran PCon on the captured data. We gauge the resulting Relm against a metrology-grade
scan as ground truth geometry. We also evaluate the (single, unified) Relm against NeRO
and RT-Splatting reconstructions in the following regards:
\begin{itemize}
  \item \textbf{Dimensional accuracy} -- Fine-grained surface accuracy gauged against
        the ground truth scan, versus NeRO~\citep{liu2023nero} (which is optimized for
        global geometric accuracy)
  \item \textbf{Perceptual accuracy} -- Fidelity of re-rendered training views, versus
        RT-Splatting~\citep{shi2026rtsplatting} (which is optimized for visual
        perception accuracy)
  \item \textbf{High-fidelity PBR mesh export} -- Geometric accuracy of the exported,
        engine-ready, relightable mesh, versus mesh exports of the NeRO and
        RT-Splatting reconstructions
\end{itemize}

\section{GSR Opportunity}
\label{sec:opportunity}

Generalized Scene Reconstruction enables easy capture and sharing of high-fidelity
immersive models of real objects and scenes. Important GSR use cases include
visualization, digital content creation, digital content streaming, mass customization,
spatial awareness, automated visual inspection, spatial AI training \& inference,
extended reality (XR), precision navigation, and disentangled
sensing\footnote{Disentangled sensing means co-locating GSR processors with image sensors
to allow the fundamental elements of a scene model to be streamed to client processes,
rather than the images, which are secondary elements.}. A forecast of total addressable
market (TAM) size for GSR is \$92B by 2031\footnote{Ask Quidient for a copy of this
forecast which is a synthesis of other market research data.}.

When forecasting the nature of emerging markets, it is often helpful to study comparable
markets. Global Positioning System (GPS) is comparable to GSR in important ways,
including:
\begin{itemize}
  \item Both GPS and GSR are measurement technologies at their core. GPS measures (to
        on-demand accuracy) locations (2D) on the surface of the earth. GSR measures (to
        on-demand accuracy) light and matter in regions (5D) comprising the earth.
  \item Both GPS and GSR require software ecosystems in which a relatively small number of
        API-First companies deliver sophisticated results used by a relatively large
        number (thousands) of software companies to deliver consumer applications.
  \item Finally, both GPS and GSR are quintessentially life-enhancing products.
        Applications critically enhanced by GPS include Google Maps, Uber, and Life360.
        Applications that will be critically enhanced by GSR include mass
        customization\footnote{\eg, ordering curtains by pointing a smartphone at a
        window.}, democratized inspection\footnote{\eg, taking care of car damage from the
        privacy of your own home.}, and persistent reality\footnote{\eg, updating a
        digital model of your home and its contents as you live in it.}.
\end{itemize}
Coincidentally, the TAMs for GPS and GSR are likely to be about the same size: \$100B.

\section{Background}
\label{sec:background}

The physical world is not made of polygon mesh surfaces and pixels. It is made of
\emph{matter} and \emph{light} that flows through and interacts\footnote{Light-matter
interaction includes reflection, transmission, refraction, scattering, and absorption.}
with it. A chrome bumper, a pane of glass, a glossy painted vase, a potted plant, the view
through a window---these are not edge cases. They are the ordinary content of the spaces
people occupy, and they are precisely where PCon avoids the shortcomings of other GSR (let
alone conventional 3D) approaches.\footnote{We note that non-PCon approaches have tended to
be transitory in nature. They ``come and go'' as practitioners \& stakeholders become aware
of their limitations.}

PCon and other GSR approaches can be roughly categorized as follows:

\begin{center}
  \includegraphics[width=0.72\linewidth]{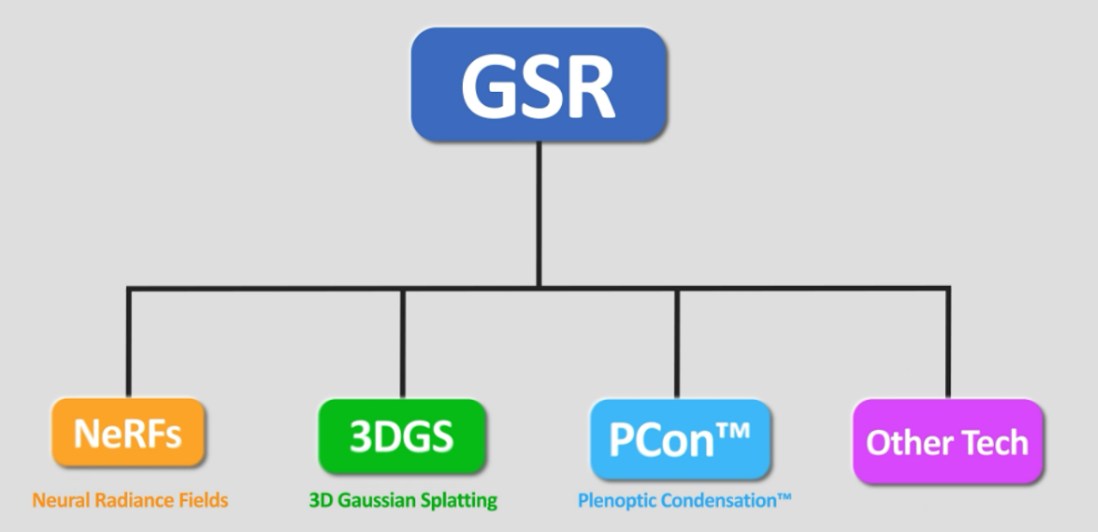}
  \captionof{figure}{Approaches to GSR}
  \label{fig:taxonomy}
\end{center}

\subsection{NeRF and 3DGS}
\label{sec:nerf-3dgs}

3D computer vision has been reshaped by the radiance field branches of GSR:
NeRFs~\citep{mildenhall2020nerf} and 3D Gaussian Splatting
(3DGS)~\citep{kerbl2023_3dgs,poirierginter2026gray,moenneloccoz2024_3dgrt}. These
approaches take in images (light field samples) of a scene and enable rendering /
synthesis of images from novel viewpoints with state-of-the-art accuracy and speed. But they
focus mainly on predicting view-dependent \emph{exitant radiance} in the original,
non-relit scene, using simplified scene representations that only weakly disentangle
light and matter (from each other)\footnote{Examples of simplifications used in such
scene representations include BSDFs that represent only isotropic opacity, emissive
(emitted) light that subsumes physically reflected or transmitted
light, and scene elements located in empty space (dry air) rather than accurately close
to the real matter they represent.}. These approaches therefore do not provide
high-accuracy relightability\footnote{Including characterization of light-interacting
media.} nor (being optimized for view synthesis) matter field geometry.

Reflection-aware variants such as Ref-NeRF~\citep{verbin2022refnerf}, the
environment-Gaussian reflections of EnvGS~\citep{xie2025envgs}, and the reflection /
transmission splatting of RT-Splatting~\citep{shi2026rtsplatting} improve the \emph{appearance}
of specular surfaces. Yet they remain render-centric. They neither relight nor produce a
usable surface mesh (for typical applications), leaving them lacking on both of the
primary scene characteristics GSR targets: geometry and relightable materials. We use
RT-Splatting as our appearance baseline (and cite EnvGS as a closely related
appearance-only method).

To summarize, NeRF and 3DGS methods suffer from the following shortfalls:
\begin{itemize}
  \item \textbf{Low relighting accuracy.} With illumination entangled into appearance, a
        radiance field cannot be accurately relit without subsequently running a
        separate inverse-rendering stage.
  \item \textbf{Low geometric accuracy.} Optimized for image synthesis, the underlying
        primitive scene elements are not accurately aligned to real physical surfaces.
        Specular highlights, which violate (simplified) photo-consistency, corrupt the
        recovered geometry.
  \item \textbf{No clean mesh.} Extracting a usable surface from millions of unorganized
        primitives is a lossy post-process, not a native output.
\end{itemize}
Plenoptic Condensation (PCon), in contrast, strongly disentangles light and matter to
overcome these shortfalls.

\subsection{Other related work}
\label{sec:related}

\textbf{Reflective surface geometry and material.} Classical SfM/MVS and neural
implicit-surface methods (NeuS~\citep{wang2021neus},
Neuralangelo~\citep{li2023neuralangelo}) recover geometry on textured, opaque,
well-behaved objects but are brittle under specularity. The reflective-object
reconstructors NeRO~\citep{liu2023nero} and TensoSDF~\citep{li2024tensosdf} jointly
recover a neural-SDF surface and a BRDF for glossy objects---NeRO via split-sum
shading~\citep{karis2013real}, and TensoSDF via a roughness-aware tensorial SDF. They are
the strongest geometry baselines for highly reflective matter, but they tend to
over-smooth fine detail and are built for reconstruction of small, statically lit
objects. We adopt NeRO as our geometry baseline (and cite TensoSDF as a closely related
method).

\textbf{Mesh and material extraction.} Delivering a usable asset requires a mesh with
materials. Radiance field and splatting methods produce meshes only as a lossy
post-process. SuGaR~\citep{guedon2024sugar} and Gaussian Opacity
Fields~\citep{yu2024gof} regularize and convert splats to surfaces.
2DGS~\citep{huang2024_2dgs} flattens its primitives for meshing---because the optimized
primitives are unorganized rather than surface-aligned. NeRO and TensoSDF do output a
marching-cubes mesh, but material recovery is a separate stage. In contrast, PCon treats
the relightable, material-bearing matter field as the \emph{native} reconstruction target, so a
PBR mesh is a low-loss export, not a conversion (Secs.~4.6 and 4.10).

\textbf{Inverse rendering and relighting.} A further line of work recovers materials and
lighting so a scene can be relit: TensoIR~\citep{jin2023tensoir},
GS-IR~\citep{liang2024gsir}, Relightable 3D Gaussians~\citep{gao2024relightable}, and
IRGS~\citep{gu2025irgs}. PCon recovers a relightable material model as part of its
unified reconstruction process. We treat relighting qualitatively here---with
self-contained consistency checks---and leave a quantitative comparison
against captured relit images (\eg, against IRGS) to future work.

\subsection{PCon's position}
\label{sec:position}

The field of non-PCon GSR approaches is split. Approaches optimized for appearance render
accurate novel views of specular scenes but do not produce accurate geometry nor
relightable materials (including PBR meshes). Approaches optimized for geometry recover
reflective surfaces but lose fine detail over large surface volumes. PCon spans the split
by jointly \emph{condensing} the matter field and light field of a scene into a unified,
tractable model (Sec.~4). This yields accurate appearance (including specular
reflections), sub-millimeter geometry, and a relightable PBR mesh. We show results,
including relighting, on a highly specular vehicle body panel with shallow dents. Because
the mechanisms behind such a result---disentanglement-driven refinement and adaptive
condensation of the scene model---are modular, they also suggest a way to strengthen
existing methods, a direction we return to in Secs.~5.3 and 6.

\section{Plenoptic Condensation (PCon)}
\label{sec:method}

This section recaps the GSR formulation that PCon builds on.

\subsection{Key characteristics}
\label{sec:keychar}

Plenoptic Condensation (PCon) turns a set of camera observations into a unified scene
model representing the matter field and light field, regionally disentangled from each
other to an adaptive degree. The name is literal: on a regional basis, PCon \emph{condenses} a
lower-power\footnote{``Power'' means ``representational power'' henceforth in this
document, unless explicitly stated otherwise.}, more ``soupy'' representation of entangled
matter and light into a higher-power, ``more structured''
representation.\footnote{Recorded images of a scene (initially without even camera pose
estimates) are a ubiquitous low-power scene model that is the typical starting point for
successive condensation stages. Note that a given scene region (in the real scene) can be
represented at multiple power levels that coexist \& overlap in space. When a query is
made for scene characteristics in that region, one power level or a weighted combination
of power levels answers the query.} The higher-power condensed representation more
faithfully represents several scene characteristics of interest, including geometry and
BSDFs of matter disentangled from light, radiance\footnote{Or other equivalent
quantity(ies) representing flux in the light field.} of light disentangled from matter,
topology of disentangled matter, dynamism\footnote{Change in (values of) characteristics
as time advances (at any time scale).} in the matter field, and dynamism in the light
field.

A defining property of PCon is that the Reality Models (Relms) it produces are unified:
appearance, geometry, and physically based (PBR) materials are all direct readouts of one
model rather than the products of separate, lossy post-processing stages. This
distinguishes PCon from pipelines that entangle (bake) view-dependent exitant radiance
into scene elements and then convert to a mesh or material as a subsequent stage.

The remainder of this section covers some relevant history of PCon (Sec.~4.2), the Relm
scene representation (Secs.~4.3--4.5), heterogeneous matter field primitives that yield
accuracy on demand (Sec.~4.6), the condensation operator (Sec.~4.7), how near-field and
far-field reflections are captured (Sec.~4.8), the optimization process (Sec.~4.9),
explainable outputs (Sec.~4.10), example Relms reconstructed using PCon (Sec.~4.11), and
a brief description of the Quidient Reality\textregistered{} scene reconstruction engine
that embodies PCon.

\subsection{History}
\label{sec:history}

We call scenes that contain highly non-Lambertian (specular), partially transmissive,
``featureless''\footnote{In the sense of features useful in conventional photogrammetry.
We sometimes call these ``features of contrast''.}, and/or finely structured matter
\emph{generalized scenes}, and we call their faithful reconstruction \emph{Generalized Scene
Reconstruction}. Quidient was founded to reconstruct generalized scenes the way they
physically exist. Our thesis, set out in the original Generalized Scene Reconstruction
paper~\citep{leffingwell2018gsr}, is that a scene should be reconstructed as a \emph{relightable
matter field} (the geometry and material response of the stuff that light interacts with)
disentangled from a \emph{generalized light field} (the radiance flowing in directions of
interest at points of interest). That first GSR paper by Quidient was lean on detail, but
it made the argument reasonably well for ``5D is the new 3D''. It describes the
light/matter formulation, practical BSDF modeling\footnote{The term ``BLIF''
(Bidirectional Light Interaction Function) is used in that paper as a synonym for BSDF.
The paper's results demonstrate optimization of BSDF parameters under a generalized light
field for the case of opaque automobile surface BRDFs.}, a multiresolution spatial
database, and reconstruction results for dented glossy automobile panels under fairly
gentle / diffuse lighting. Image data capture was performed with an imaging
polarimeter\footnote{A camera capable of recording characteristics of the polarization
state of light, in addition to its intensity (radiance) and color.} mounted on a
motorized pan/tilt head on a sturdy tripod. Coded optical targets (ArUco tags) were
placed directly on the panel surfaces to aid surface geometry initialization and camera
pose estimation.

This technical note presents the next generation of that work, including reconstruction
results that are similar in nature but much more in-the-wild, including typical indoor
ceiling lighting (highly anisotropic), handheld capture with a non-polarimetric phone
camera, and the absence of optical targets (or other added photogrammetric features) on
the panel. PCon inherits the same overall GSR formulation---matter and light coupled by
\eq{eq:opt}---but revisits how the matter field and light field are represented and
reconstructed, as described below. Our perennial use of dented automobiles to advance PCon
development is highly intentional: in addition to being a valuable beachhead market for
Quidient, dented automobiles are emblematic of the generalized scene characteristics that
drive widespread demand for GSR.

\subsection{Light, matter, and the Reality Field}
\label{sec:realityfield}

Our GSR system models scenes as two separate, disentangled fields that interact with each
other. We refer to the first field as the \emph{matter field}, a collection of volumetric
media elements, denoted henceforth as mediels. At the boundary between sufficiently
distinct media, mediels specialize to the familiar surface element (surfel).

Paired to the matter field is the \emph{light field}, a collection of radiometric
elements called radiels that model (the distribution of) incident or exitant light
flowing within some solid angle (range of directions) about the radiel's point of origin.
Viewed in this manner, the incident light arriving at any point and within some solid
angle is represented as an incident radiel. The incident radiel's (distribution of)
radiance can be computed by transporting exitant radiance from the set of nearest (first
encountered) ``upstream'' mediels falling within (subtended by) the radiel's solid angle.
This generalized approach admits modeling of complex light transport scenarios, including
the presence of dynamism. The disentangled nature of the two fields also notably allows
for replacement of light fields to relight a scene.

Together, the matter field and light field form the \emph{Reality Field}. This is a
unified, 5D representation that composes the 3 dimensions of position plus the 2
dimensions of direction of the field of light. We treat the Reality Field as agnostic of
underlying data structures and light transport algorithms. It is simply an abstraction to
containerize (representations of) mass and energy. Disentangled Reality Fields form the
basis of Quidient's relightable Reality Models.

\subsection{Scene reconstruction as inverse light transport}
\label{sec:inverse}

PCon poses scene reconstruction as an optimization that drives (parameter values of) predicted radiels toward (those of) observed radiels over the degrees of freedom of the Reality Field: 
\begin{equation}\label{eq:opt}
  \min_{f_\ell,\,L}\ \sum_{(x\to\omega)\,\in\,\mathrm{obs}}
    \rho\!\left(L_{\mathrm{pred}}(x\!\to\!\omega;\, f_\ell, L) - L_{\mathrm{obs}}(x\!\to\!\omega)\right),
\end{equation}
where $\rho$ is a robust, uncertainty-weighted consistency measure. Solved over a region
of mediels, \eq{eq:opt} unifies and subsumes several classical shape-from-X objectives
(shape-from-shading, -polarization, -silhouette;
structure-from-motion).\footnote{This radiometry-driven optimization approach can, for
example, be used to disentangle a flashlight's true (via converting electrical energy to
optical electromagnetic energy) emitted light vs other incident light from the
surrounding scene that gets reflected off the flashlight's lens (and curved reflective
surface).} It is typically optimized in stages---alternating refinement of the light
field and the matter field---initialized from sparse structure-from-motion and prior
scene information.

\subsection{Inputs and native representation}
\label{sec:inputs}

PCon takes in a set of images with initial (estimates of) intrinsic and extrinsic camera
parameters, typically recovered by structure-from-motion. Light reflecting from specular
surfaces spans a high dynamic range, so PCon reconstructs HDR scene-linear radiance, which
is available using consumer mobile device cameras\footnote{Available, for example, from
Apple ProRes Log imaging on a consumer iPhone and decoded to 32-bit linear EXR with
apple-log2linear~\citep{tilmon2025applelog}, retaining unclipped highlights for
physically based light-transport fitting (while still potentially using the 8-bit sRGB
stream for pose estimation).}. High-fidelity capture is thus accessible from an ordinary
mobile device afterward. The reconstruction output is represented natively as a condensed
Reality Field (Sec.~4.3). We deliberately leave the concrete data structure open here;
PCon does not depend on a particular one. Crucially, this 5D structure is not an
intermediate scaffold to be discarded---it is the deliverable, and every output in
Sec.~4.10 is computed from it directly.

\subsection{Heterogeneous matter field elements}
\label{sec:primitives}

The matter field is represented as a \emph{heterogeneous} set of primitive elements, and this is
central to PCon's geometric accuracy. Scene regions with low geometric detail (\eg,
curvature) and/or low specularity are represented efficiently by Gaussian-like
primitives, in the spirit of point-based rasterization~\citep{kerbl2023_3dgs}. Where the
geometric or radiometric residual is high (\eg, sharp edges, fine structure, gaps between
surfaces, and high-frequency specular response), those primitives are \emph{condensed on demand}
to higher-order primitives (oriented curve, surface, and volume elements offering more
physically truthful representations of geometry, topology, and light interaction) that
resolve detail a Gaussian scene element cannot.

This condensation is typically driven by regional exitant radiance loss (residuals) and
proceeds coarse-to-fine, so representational power (and compute) is spent only where the
reconstruction goal requires it. The effect is accurate sub-millimeter geometric detail
in the regions that need it, without the cost of globally dense high-power
primitives---and a graceful accuracy / efficiency tradeoff.
This mixed-primitive and mixed-power strategy is, to our knowledge, unique to PCon. It is
also what a single neural-SDF surface cannot do. One global field of fixed representational
power must be spent everywhere at once and averages out fine structure when applied to
larger reconstruction volumes (\eg, car or house). In contrast, error-driven condensation
is capacity on demand---it scales to the whole reconstruction volume while still resolving
sub-millimeter detail, but only where the regional loss calls for it.

\subsection{The condensation operator}
\label{sec:condensation}

Each surface location is observed by many cameras, each measuring (the radiance of) a
single radiel exitant from the location. PCon condenses these per-view measurements into
the compact per-mediel geometry and BSDF state that explains them. For a mediel $x$, the
observed exitant radiels $\{L_{\mathrm{obs}}(x\to\omega_i)\}$ are aggregated into an
exitant light field; the incident light field of incident radiels
$\{L_{\mathrm{pred}}(x\leftarrow\omega_j')\}$ is assembled by transporting light that is
exitant from the surrounding ``upstream'' scene elements (Sec.~4.3); and a BSDF is
fit so that the (predicted) incident light field reproduces the (observed) exitant light field
under the light-transport relation. The redundancy across views is what makes
the incident and exitant light fields \emph{compressible} into a small number of matter (and
emissive light) parameters---hence ``condensation''.

\paragraph{Why condensation suits highly specular matter.} For a highly specular surfel, a
single exitant radiel constrains only the product of the BSDF and the incident radiance
along one direction (very narrow solid angle). Many exitant viewpoints over a non-trivial,
reconstructed incident light field over-determine the BSDF and thereby strongly
disambiguate geometry from appearance (including in a highly specular regime where
radiance-baking methods cannot).

\subsection{Near-field and far-field reflections}
\label{sec:reflections}

Reflections carry information from multiple ranges, which can be roughly categorized into
\emph{near-field} and \emph{far-field}. PCon represents both within the single Reality Field. Far-field
reflections---the sky, distant buildings, and foliage mirrored in a car's paint---are
modeled well by a shared, global far-field light field queried by direction alone.
Near-field reflections---the surrounding room, the user (or, \eg, UAV) holding the camera,
nearby objects---vary with position and require spatially varying incident light fields,
which PCon obtains by light hopping between mediels. Because PCon's light field
representation is multi-resolution in direction space, angular resolution is allocated
where reflections are sharp, and coarsened elsewhere. This is why a single PCon scene
model reconstructs and resamples reflections both indoors (near-field dominant) and
outdoors (far-field dominant).

\paragraph{Modular light-field backing.} Because PCon's incident light field model is
modular, its initialization and refinement can be driven by direct camera observations
(image pixels) of the incident light and/or by optimizing for BSDF consistency at scene
elements receiving the incident light, following image-based lighting principles. When the
scene environment is under-observed, one or more provided environment maps (IBL) or a
learned indirect representation could also be used. Because every variant enters
through the incident field rather than through the matter field, swapping
between them leaves the recovered BSDFs and geometry untouched. This modularity also
naturally connects to relighting: a relighting environment map is itself image-based
lighting swapped in for the recovered original incident field.

\subsection{Optimization}
\label{sec:optimization}

PCon solves the inverse-light-transport objective of \eq{eq:opt} by \emph{alternating refinement}:
holding the matter field fixed while the light field is updated, then holding the light
field fixed while mediel geometry, power level, and BSDFs are updated, under robust,
uncertainty-weighted residuals. Optimization is multi-resolution---regionally adaptive
coarse-to-fine refinement of the Reality Field---with condensation of scene elements
(Sec.~4.7) triggered by local residuals. The overall optimization problem is typically
realized as several smaller optimization sub-problems that are adaptively formulated
across different scene regions in (position) space and time. The subproblem regions can
overlap (in time as well as space). The overall optimization problem is typically
initialized from sparse SfM and a priori scene structure. Camera poses may be refined
jointly with other Reality Field parameters.

\subsection{Explainable outputs: appearance, geometry, and PBR mesh}
\label{sec:outputs}

Because the reconstructed Reality Field is disentangled / relightable, the capabilities
evaluated in Sec.~5 use \emph{readouts} of one model, not separate reconstructions:
\begin{itemize}
  \item \textbf{Appearance.} Transport the light exitant from a Reality Field region to a
        virtual camera to render an image (Sec.~5.3).
  \item \textbf{Geometry.} Read the spatial configuration (including topology and
        topography) of surface, curve, point, and volume structures in the matter field,
        with sub-millimeter detail in highly condensed regions, as a mesh or point set
        (cloud) for metric comparison against the structured-light scan (Sec.~5.3).
  \item \textbf{High-fidelity PBR mesh.} Export a watertight mesh whose fitted BSDFs are
        written directly to standard physically based rendering (PBR) material
        channels---base color, roughness, metallic, and normals---together with the
        recovered environment lighting. Because materials and lighting are reconstructed
        natively rather than approximated after the fact, this is a \emph{low-loss export} (an
        engine-ready, relightable asset), not a lossy bake; we quantify its geometric
        accuracy, and the small loss relative to the native reconstruction, in Sec.~5.3.
  \item \textbf{Relighting.} Replace the recovered incident light field with a new one and
        re-evaluate radiance seen by the camera through the unchanged BSDFs to render under
        novel illumination. A quantitative benchmark against captured relit images is future
        work (Sec.~6).
\end{itemize}
No competing baseline supplies appearance, sub-millimeter geometry, \emph{and} a relightable PBR
mesh from one model. PCon does.

\subsection{Examples}
\label{sec:examples}

In this subsection, we share example reconstructions using PCon.

We showcase important aspects of PCon using reconstructions of a mirrorlike iconic
``balloon dog'' and a boxing glove with multiple surface materials of different sheens:
\begin{center}
  \includegraphics[width=0.72\linewidth]{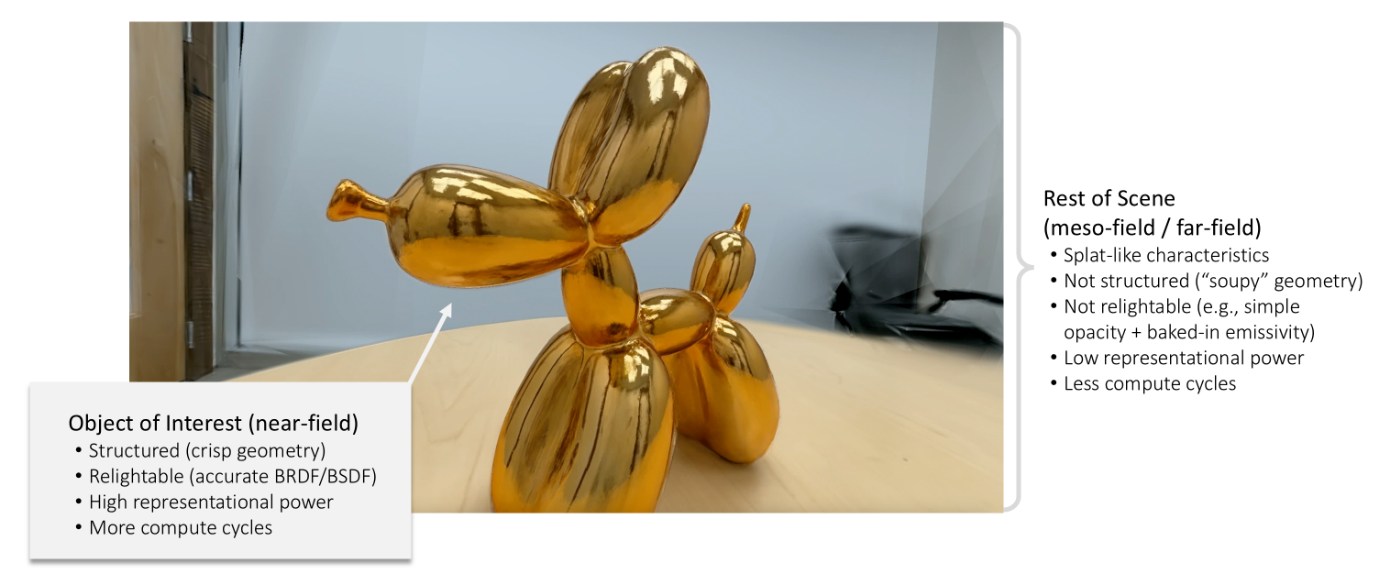}
  \captionof{figure}{Adaptive representational power across scene regions}
  \label{fig:adaptive}
\end{center}
\begin{center}
  \includegraphics[width=0.82\linewidth]{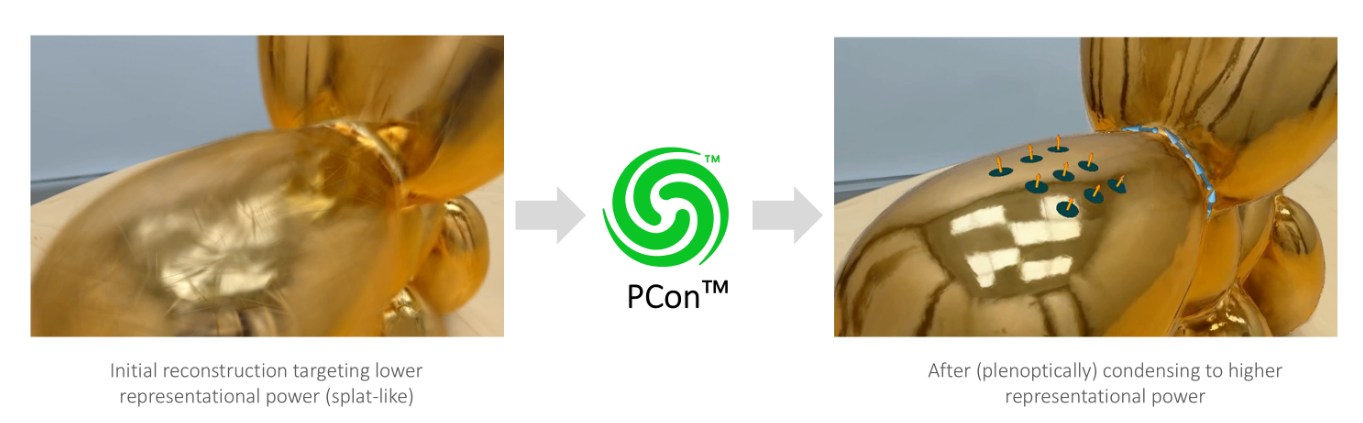}
  \captionof{figure}{Plenoptic condensation from a lower-power ``soupy surface'' to a
    higher-power structured surface (application of the condensation operator)}
  \label{fig:condensation}
\end{center}

\noindent Relighting with comparison to SOTA non-relightable 3D Gaussian Splatting (3DGS):
\begin{center}
  \includegraphics[width=0.82\linewidth]{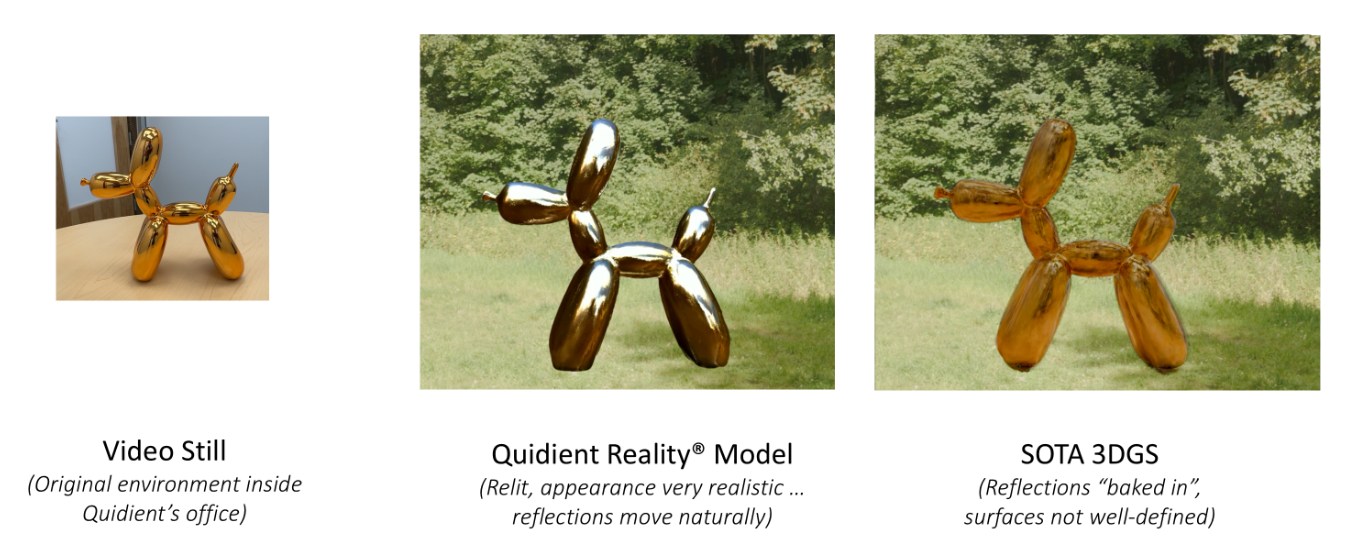}
  \captionof{figure}{Relighting with comparison to SOTA non-relightable 3D Gaussian
    Splatting (3DGS)}
  \label{fig:relight3dgs}
\end{center}
\begin{center}
  \includegraphics[width=0.62\linewidth]{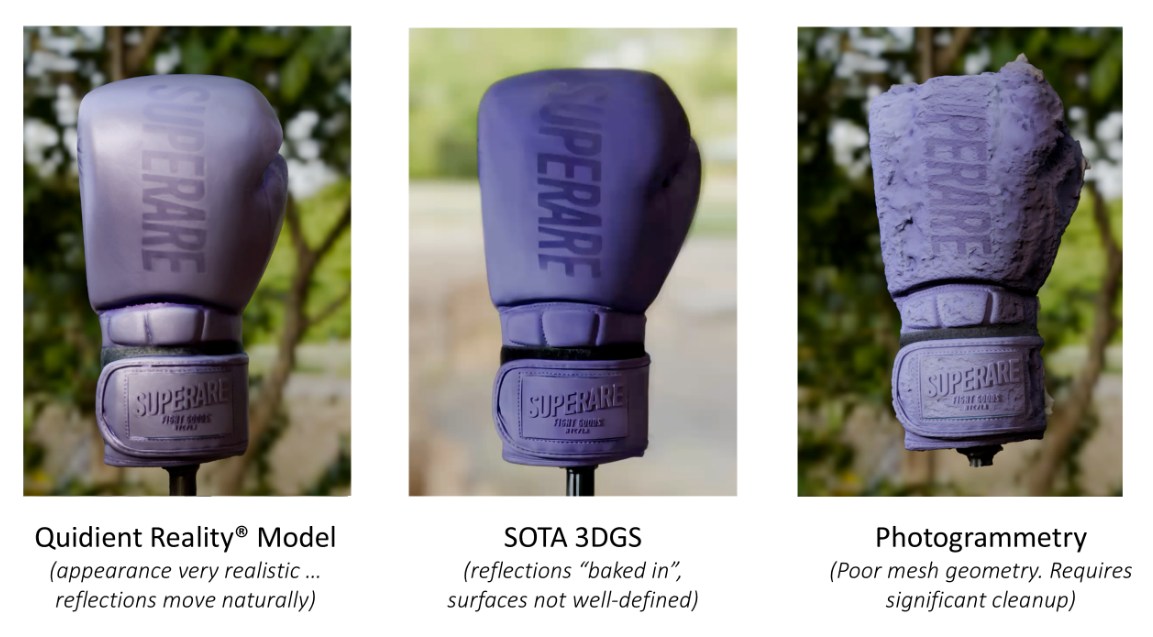}
  \captionof{figure}{Relighting with comparison to SOTA non-relightable 3DGS and
    conventional photogrammetry}
  \label{fig:relightglove}
\end{center}

\noindent The following mosaic shows PCon Reality Models reconstructed from captures done
with phone and drone (UAV) cameras (none of these are photographs; they are all renders of
in-the-wild reconstructions):
{\centering
  \includegraphics[width=\linewidth]{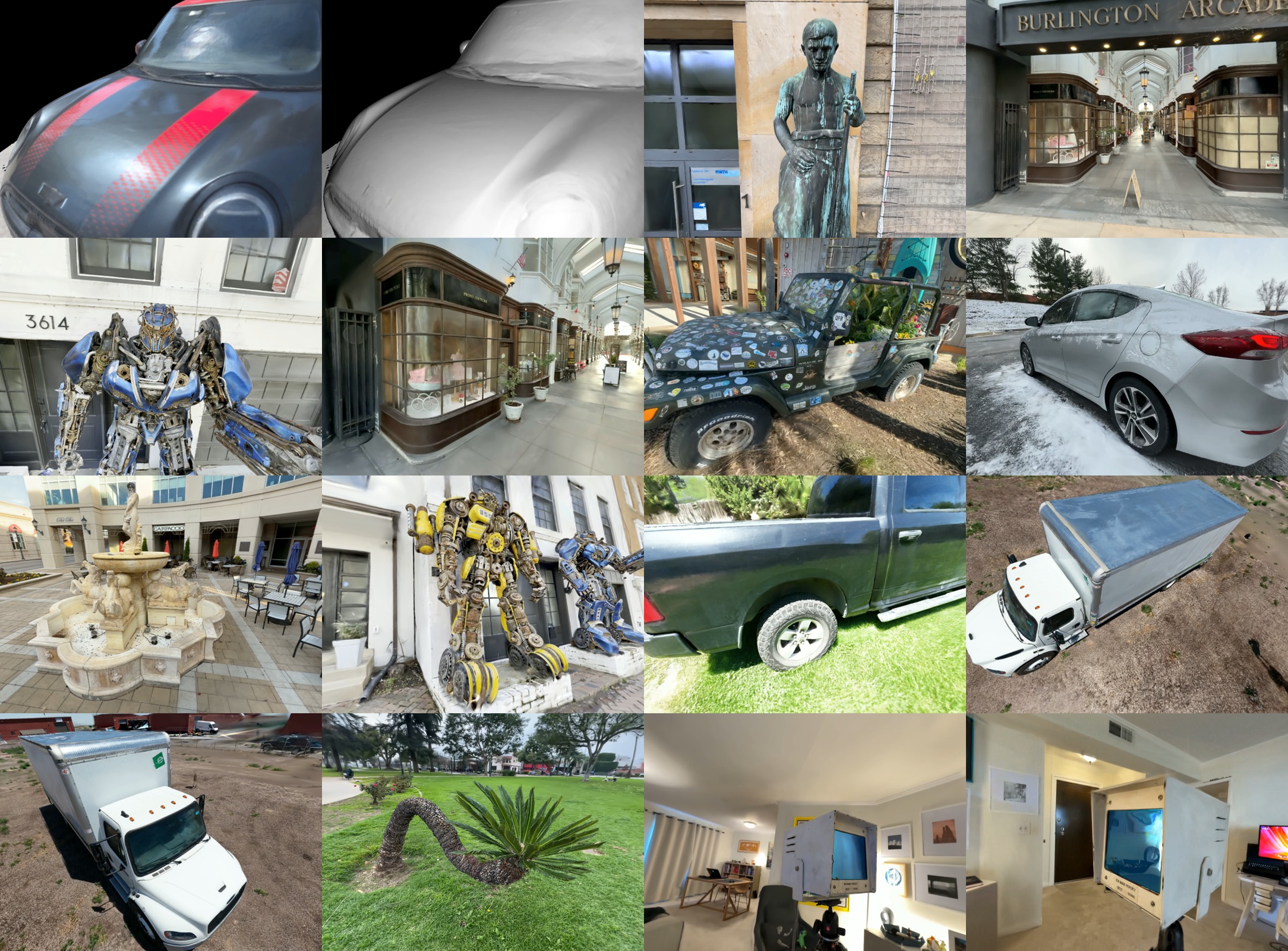}\par}
\vspace{1mm}
\captionof{figure}{\textbf{Not} photographs. In-the-wild PCon reconstruction renders,
  casually captured by smartphones and drones. These reconstructions are from varying power
  levels of PCon, meaning some reconstructions show higher-power light transport modeling
  such as reflections, while other reconstructions show high resolution but lower-power
  light transport modeling in the near and far field as needed by PCon.}
\label{fig:examples}

\subsection{Quidient Reality\textregistered{} Engine}
\label{sec:engine}

PCon is core to the Quidient Reality\textregistered{} Engine, which is being used by
Quidient development partners to develop remarkable products. Engine capabilities include:
\begin{itemize}
  \item \textbf{Mainstream GSR database} -- Enables efficient query and updating of
        persistent Reality Models (Relms) that represent scenes from the size of a room to
        the size of the earth.
  \item \textbf{Generalized mapping} -- Camera pose estimates are refined jointly with the
        matter field and light field, reducing reliance on photogrammetric features of
        contrast (including engineered optical targets). Quidient Reality is able to
        perform high-fidelity reconstructions in the wild.
  \item \textbf{Adaptive fidelity} -- Representational power\footnote{Simply ``power'' as
        shorthand.}, accuracy, and resolution vary across scene regions to meet
        reconstruction goals while honoring constraints like compute budget.
  \item \textbf{Uncertainty tracking} -- The physical truthfulness of reconstructed scene
        parameters is quantitatively tracked and queryable at will. This serves as a
        robust confidence metric.
  \item \textbf{Intuitive capture} -- Scene capture that's easy enough to be done by
        ``granddads.'' Capture (including guidance to the user) is driven by reconstruction
        goals.
  \item \textbf{Subscenes (and superscenes)} -- Subscenes can be extracted, composed,
        processed, and reconciled with a larger containing scene (superscene). Enables
        reconstruction of boundless scenes with nearly unlimited detail. Supports efficient
        scene streaming and updating / maintenance of large scene models (hosted in the
        cloud). Using ``intelligent dispatch'', extracted subscenes can be routed to
        CPU/GPU processing nodes to honor latency and cost margins while meeting processing
        goals.
  \item \textbf{Segmentation} -- Reality Models can be segmented in an ``explainable'' way
        according to their various characteristics (including geometry, BSDF, light field
        pattern, topology, temporal dynamism). Essential for scene understanding, a given
        Reality Model can be segmented into multiple ``segtrees'', each regarding different
        sets of characteristics.
  \item \textbf{Training and inference on Relms} -- Reconstructed Reality Model fidelity
        is high enough to support ML training and inference directly on native scene
        elements. Inference performance can be equivalent or better with less Relm training
        data versus 2D image training data.
  \item \textbf{Strong dynamism modeling (in future v2.0)} -- Significant temporal change
        in matter, light, and spatial (coordinate) frames of reference is represented in
        Reality Models. One notable capability afforded by strong modeling of scene
        dynamism (which is another type of disentanglement) is to render images at novel
        times (and viewpoints), as opposed to merely novel viewpoints.
\end{itemize}
PCon (Relms) is already being tested by Fortune 100 development partners in important
commercial, defense, and aerospace inspection applications.

\section{Experiments}
\label{sec:experiments}

We evaluate PCon on a real Fiat 500 automobile in our imaging lab. We show results on a
higher power level condensed Reality Model that has advanced disentanglement needed for
accurate geometry and appearance reconstruction on specular surfaces. Relighting is
supported, but a quantitative relighting benchmark against captured relit images is left
to future work. We choose to compare NeRO as a neural signed distance function
representation, being loosely NeRF-adjacent, since it is academically SOTA for geometry
and to a lesser extent appearance on reflective surfaces due to its neural field-based
light transport modeling. We choose to compare RT-Splatting, being 3DGS-adjacent, as it
is academically SOTA for appearance on reflective and transmissive scenes, while showing
no geometry metrics in their paper besides qualitatively convincing surface normal and
depth renders that imply decent global geometry reconstruction, but not necessarily highly
multiview-consistent geometry needed for quality local geometry reconstruction and mesh
extraction.

\subsection{Dataset and capture}
\label{sec:dataset}

\paragraph{Capture.} The Fiat 500 is imaged from an iPhone 17 Pro in Apple ProRes Log 422LT
format ensuring favorable settings for metadata-based radiometrically and geometrically
calibrated video with minimal compression artifacts. We convert the log video to sRGB,
ensuring compatibility with competing methods' color space assumptions, while noting that
PCon supports and improves with HDR RAW images when available~\citep{tilmon2025applelog}. We
ensure the light field incident to the car and the light field exitant from the car are
observed from a sufficiently diverse set of camera viewpoints. We use 300 keyframes for
reconstruction as determined automatically from our reconstruction system. No markers are
used to aid pose estimation, but we do place feature-rich posters on the wall.

\begin{figure}[htbp]
  \centering
  \includegraphics[width=\linewidth]{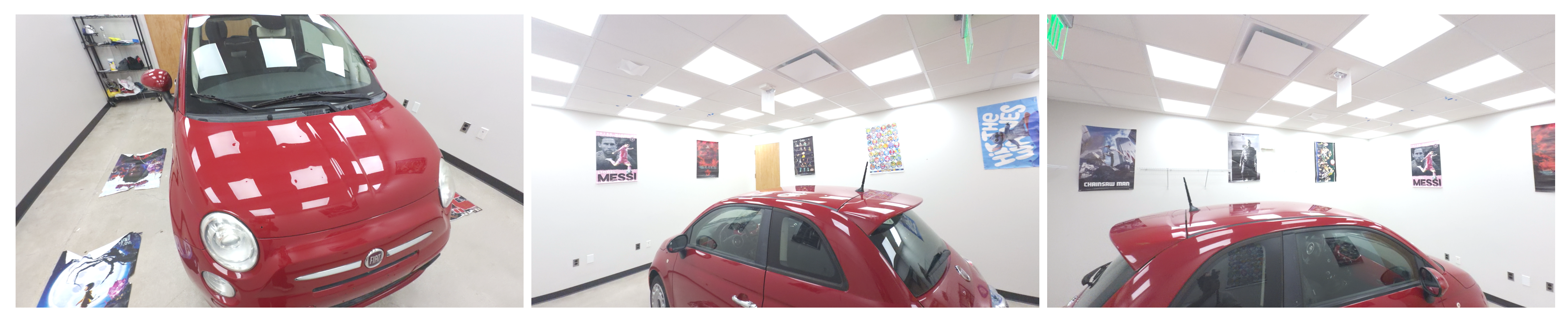}
  \caption{Example images from the dataset captured in our imaging lab}
  \label{fig:dataset}
\end{figure}

\begin{figure}[htbp]
  \centering
  \includegraphics[width=0.62\linewidth]{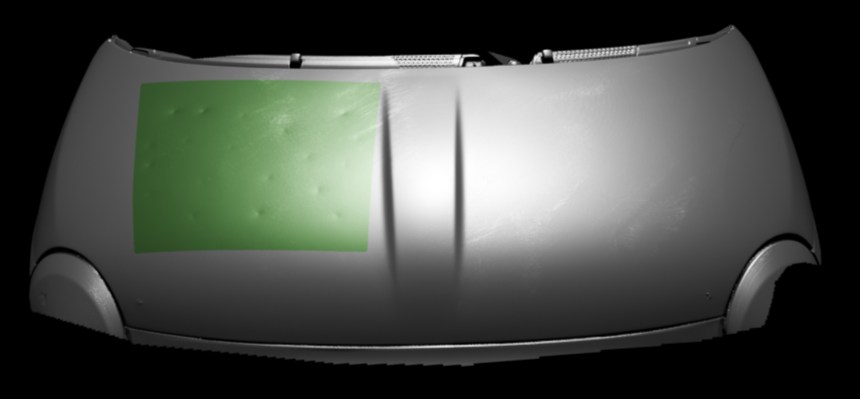}
  \caption{Ground truth Artec Spider II structured light scanned geometry of the Fiat hood
    used for GT geometry and relighting comparisons. The evaluated region is shown in green
    on the left half of the hood, on the dent cluster, with follow-up results showing the
    individual dent geometry profiles. In the Table 1 results, the green patch is the
    global geometry for which chamfer distance is computed. The local dent profile Mean
    Absolute Error geometry results are for the individual dents within this green patch.
    The noise on the surface is due to the scanner picking up residue from the matte
    spray coating that must be applied to reflective surfaces for accurate scanning.}
  \label{fig:scan}
\end{figure}

\paragraph{Ground-truth geometry.} A metrology-grade structured light scan from an Artec
Spider II is used as the GT surface (in mesh form) for all comparisons. We evaluate global
geometric accuracy within a manually defined patch region (shown in green in
Figure~\ref{fig:scan}), and local damage profile geometric accuracy on manually defined dent
regions, mirroring the regional accuracy assessment of the original GSR
study~\citep{leffingwell2018gsr}. Due to structured light scanners' inability to work well
on specular surfaces, the vehicle's hood is spray-coated with a matte spray before scanning.

\subsection{Protocol}
\label{sec:protocol}

\paragraph{Metrics.} We quantify the appearance, geometry, and time of GSR approaches in
the following manner. Appearance metrics are presented with PSNR, SSIM, and LPIPS from RGB
renders of the reconstruction compared to a masked region of the ground truth. We use Chamfer
Distance and normal vector Mean Angular Error to categorize the global patch accuracy and
mesh vertex Mean Absolute Error for individual dent damage categorization. Additionally,
we track reconstruction and rendering time for all methods.

\paragraph{Baselines.} We employ RT-Splatting~\citep{shi2026rtsplatting} as a modern
representative of predominantly appearance-oriented reconstruction, and
NeRO~\citep{liu2023nero} for a geometry-centered technique. For both methods, we evaluate
geometry on a final extracted mesh via marching cubes. In the case of RT-Splatting, this
mesh is acquired from a \emph{depth-fusion proxy}: we fuse its rendered depth over all training
views (masked with our GT patch hit mask from aligned GT) into a point cloud and compare
it to the scan under the same region, alignment, and sampling. We report this proxy
because it quantifies the geometry an appearance-oriented renderer can supply. However, we
note that rendered depth on specular paint is often erroneous, especially for 3DGS-based
methods, so the large reported error reflects a genuine limitation of the methodology.

\paragraph{Fairness.} To ensure an impartial test of all GSR methodologies, each approach
is initialized with the same set of initial camera poses, intrinsics, and training images.
The bounding sphere for NeRO is manually defined to tightly fit the car hood to maximize
representational capacity of their neural fields. The neural angular far field light field
assumption (direction only, aka 2D) used in RT-Splatting is insufficient for the near field
imaging lab we capture in, but we view this as a limitation of their method compared to
PCon, which supports near (4D/5D) and far light field reconstruction and modeling. NeRO
supports near and far field light field reconstruction as well, albeit at low resolution with
low explainability.

\paragraph{Geometric evaluation.} We compare the Chamfer Distance and Normal Mean Angular
Error of the extracted 0.5m\textsuperscript{2} patch region meshes from PCon, NeRO, and
RT-Splatting to the same patch registered from the GT structured light scan. We register
each method's full hood mesh to the ground-truth scan via rigid Iterative Closest Point (ICP)
alignment. After registering the full hood for each mesh, we crop the patch from them and do
not further align the patch. From these registered patches, we compute the Chamfer and N-MAE
metrics against the GT of each scan (Table~\ref{tab:main}). After the full patch is extracted,
we separately crop each dent out of the patch. The visual dent curvature was used as a
reference to get dent patches. In regions where the dent is not visible or present, the dent
was located in the GT mesh, and then the mesh to GT alignment is used to get the closest patch
for the mesh under test. At this scale, ICP is no longer useful for alignment, so the
individual dents are realigned to the GT by first centering each crop in the XY-plane, then
mean-centering the dent profile along the depth (Z-axis). We then replot the dent profile's
Z-deviation against a planar fit to the cropped dent region's outer rim. This process is
analogous to the moving least squares surface projection algorithm, which removes any angular
orientation difference between GT and reconstructed dent profiles. The replotted profiles are
guaranteed to be perpendicular to the XY-plane. We quantify the dent profile accuracy by
computing the mean absolute distance between a fourth order polynomial fit of points sampled
on the reconstruction and a similar polynomial fit to the corresponding GT crop region.

\paragraph{Appearance evaluation.} While most differentiable and inverse rendering papers
employ a hold-out validation strategy to test photometric accuracy, PCon's primary
use case is not well served by that approach. Specular reflections from dents on car panels
are often only visible from a narrow region of solid angle, often around grazing angles.
The real-world capture procedure makes it difficult to gather enough distinct viewpoints
to both reconstruct the dents and hold out a set for novel-view synthesis. Instead, we
choose to compare views used for training against real-world captured images, since the
ability to accurately reproduce patterns of specular highlights is a solid indicator that
a dent's shape is accurately represented in the mesh topography. Aside from the differences
in our image selection strategy, the appearance metrics for reconstructed image synthesis
and relighting are computed in a standard manner with PSNR, SSIM, and LPIPS.

\subsection{Results}
\label{sec:results}

Table~\ref{tab:main} is the main result: one PCon Reality Model, evaluated for both appearance and
geometry against the two SOTA methods. PCon is reported at a higher (more strongly
condensed) power level that achieves advanced disentanglement.

\clearpage
{\centering
  \resizebox{\linewidth}{!}{%
    \begin{tabular}{l c ccc ccc cc}
        \toprule
        & \multicolumn{1}{c}{\textbf{Relightable}} & \multicolumn{3}{c}{\textbf{Appearance}} & \multicolumn{3}{c}{\textbf{Geometry}} & \multicolumn{2}{c}{\textbf{Time}} \\
        \cmidrule(lr){3-5}\cmidrule(lr){6-8}\cmidrule(lr){9-10}
        \textbf{Method}
          &
          & PSNR\,$\uparrow$ & SSIM\,$\uparrow$ & LPIPS\,$\downarrow$
          & Chamfer [global] (mm)\,$\downarrow$ & Damage Profile MAE (mm)\,$\downarrow$ & Normal MAE  ($^{\circ}$)\,$\downarrow$
          & Recon (hr) $\downarrow$ & Render (fps) $\uparrow$ \\
        \midrule
        PCon (PL3)   & \cmark & \cellfirst{38.94} & \cellfirst{0.9939} & \cellfirst{0.0212} & \cellfirst{2.19\,$\pm$\,0.05} & \cellfirst{0.034\,$\pm$\,0.005} & \cellfirst{1.08} & \cellfirst{0.68}   & \cellfirst{60}  \\
        RT-Splatting & \xmark & \cellsecond{37.45}  & \cellsecond{0.9936}  & \cellsecond{0.0226}  & \cellsecond{4.36\,$\pm$\,0.05} & \cellthird{0.481*\,$\pm$\,0.005} & \cellsecond{6.68} & \cellsecond{1.4} & \cellsecond{33.28} \\
        NeRO         & \cmark & \cellthird{23.88}  & \cellthird{0.9768}  & \cellthird{0.0500}  & \cellthird{6.64\,$\pm$\,0.05} & \cellsecond{0.094*\,$\pm$\,0.005} & \cellthird{6.82} & \cellthird{7}   & \cellthird{0.016} \\
        \bottomrule
    \end{tabular}}\par}
\vspace{1mm}
\captionof{table}{\textbf{Main results on the Fiat.} A single PCon Reality Model, evaluated for
    \emph{appearance} (PSNR/SSIM/LPIPS on all views used for reconstruction), \emph{mesh
    geometry} (Chamfer distance and surface-normal MAE vs.\ the structured-light scan),
    \emph{time} (reconstruction and rendering), and whether the recovered output is
    \emph{relightable}. PCon achieves the best appearance and significantly better
    geometry results among the three methods, which is significant when compared to
    surface-based reconstruction methods such as NeRO that use a neural signed distance
    function representation with surface-oriented neural light transport modeling. While
    NeRO's mesh export is relightable, their native rendering is slow due to sphere
    tracing, ray marching backpropagation, and using unoptimized MLP neural fields for all
    light transport modeling. An asterisk (*) on the Damage MAE column indicates that,
    because the dent depths are typically only around 100 $\mu$m, NeRO and RT-Splatting
    entirely fail to reconstruct most dents, with NeRO achieving only slight damage
    reconstruction for the biggest dents.}
  \label{tab:main}

\vspace{2mm}
{\centering
  \includegraphics[width=0.72\linewidth]{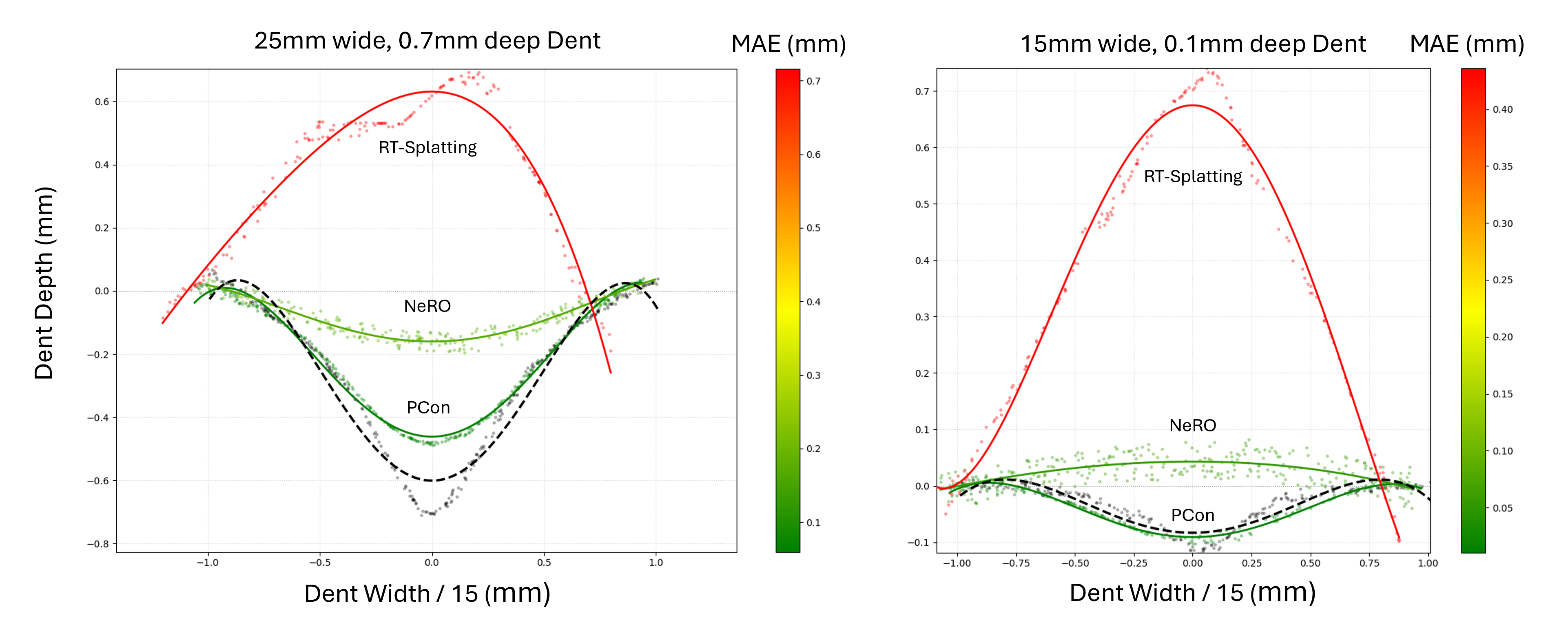}\par}
\vspace{1mm}
\captionof{figure}{We compare the dent damage Mean Absolute Error of the mesh vertex profile of
    PCon, NeRO, and RT-Splatting to the GT dent geometry from the Artec structured light
    scanner. We find that PCon achieves excellent reconstruction of the dent geometry
    accurate to within 100$\mu$m on the 25mm wide and 0.7mm deep dent, and within 6$\mu$m on the
    15mm wide and 0.1mm deep dent where NeRO and RT-Splatting completely fail to
    reconstruct the dent geometry due to how subtle and shallow the dents are. The GT
    geometry is a black dashed line. Please see Table 1 for the Damage MAE averaged over all
    19 dents on the Fiat hood patch.}
  \label{fig:dent}

\vspace{2mm}
{\centering
  \includegraphics[width=0.5\linewidth]{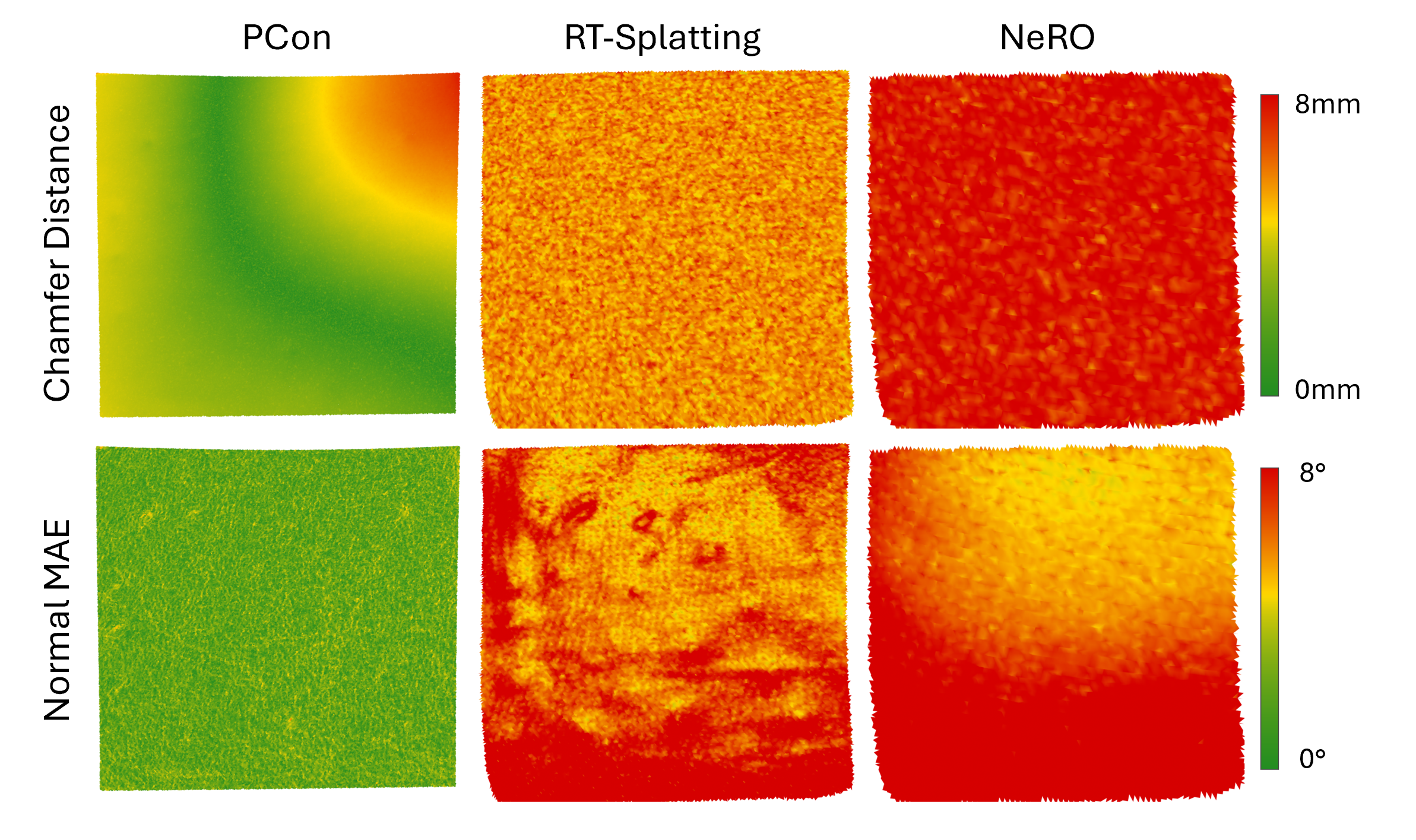}\par}
\vspace{1mm}
\captionof{figure}{Mesh Chamfer Distance and Normal Mean Angular Error heatmaps of PCon,
    RT-Splatting, and NeRO over the global mesh patch compared to the GT mesh patch after
    whole hood best fit adjustment. We note the full hood mesh from each method was used to
    independently ICP align each method's reconstruction to GT before cropping to the patch
    shown, and no further alignment was done to the patch after cropping.}
  \label{fig:heatmap}

\begin{figure}[t]
  \centering
  \begin{minipage}{\linewidth}
    \includegraphics[width=\linewidth]{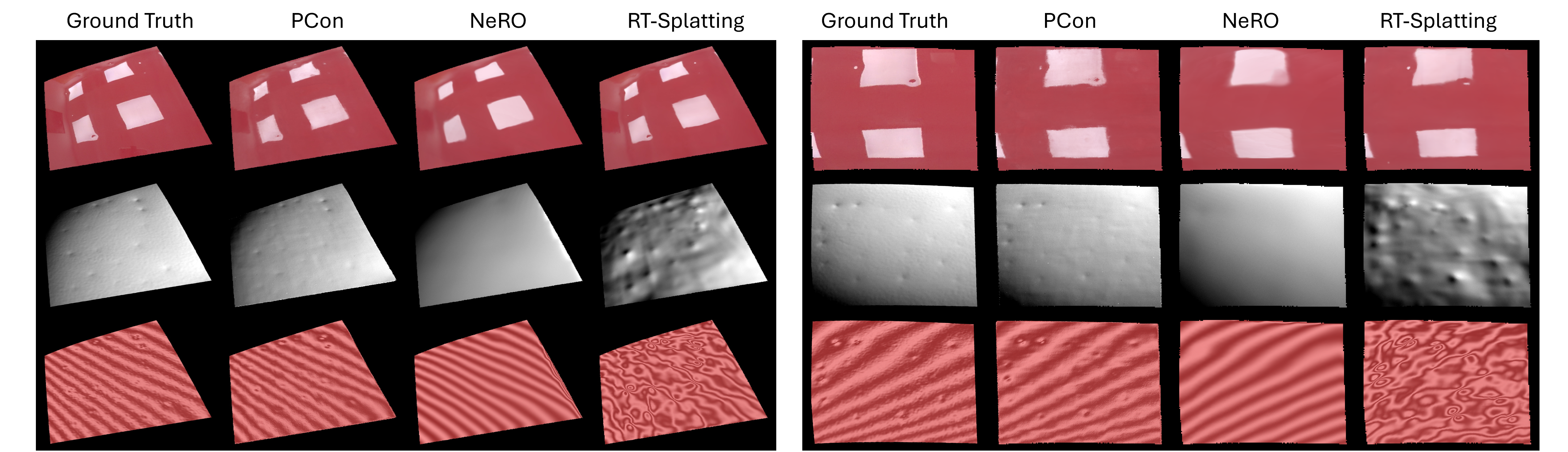}
  \end{minipage}
  \caption{More views like Figure~\ref{fig:teaser} qualitatively comparing appearance and
    geometry of PCon to NeRO and RT-Splatting}
  \label{fig:morecompare}
\end{figure}

\newpage
\paragraph{Appearance.} RT-Splatting shows impressive appearance reconstruction,
especially considering how specularly reflective the car hood surface is. However, its
geometry extraction and metrics reveal highly erroneous geometry reconstruction, indicating
that its appearance-oriented light transport is insufficient for damage measurement
applications. NeRO also achieves impressive appearance reconstruction, and while it indeed
uses geometry-oriented light transport modeling, the limited capacity of its neural fields
shows that the damage profile is over-smoothed and not reconstructed in the RGB renders. PCon
includes the strength of both methods, resulting in better appearance reconstruction that can
show the damage reflecting the incident light field in a physically based manner.
Figures~\ref{fig:teaser} and \ref{fig:morecompare} show appearance renders in the first row.

\paragraph{Geometry.} RT-Splatting geometry results indicate that its appearance results
are not due to strong local geometry reconstruction and are instead due to moderately
accurate global geometry reconstruction that is heavily supplemented with neural field
modeling of a simplified BSDF and incident light field that ensures smooth optimization and
reflections without requiring accurate local geometry, which indeed may suffice for pure view
synthesis applications. NeRO's neural SDF and other neural fields must spend representation
capacity everywhere at once, so detail is averaged out at vehicle scale. NeRO produces output
geometry that is over-smoothed and heavily bias-offset, presumably from insufficient accuracy
in light transport modeling. PCon adaptively condenses between power levels, promoting
higher-order primitives and supersampling; this is capacity-on-demand rather than a globally
uniform representation. This modeling results in better global geometric accuracy of the patch
and better local geometric accuracy of the damage that can impressively recover the physical
profile of sub-mm deep dents in a highly reflective surface.

\paragraph{PCon as an enhancer (forward-looking).} The two ingredients behind these
results are modular. The first is better matter-field modeling from adaptive condensation,
and disentanglement-driven geometry that ties the surface to the measured incident and exitant
light fields. The second is better, modular light-field modeling. We did not demonstrate this
on the Fiat, but in principle each ingredient could make an existing method more efficient and
accurate: an appearance/splatting renderer such as RT-Splatting could gain
disentanglement-driven geometry refinement from PCon, and a neural-SDF method such as NeRO
could gain adaptive condensation and a richer measured light field for finer geometry and
light field reconstruction and modeling. We treat this as a direction rather than a result,
and we hope the academic community and industry will consider PCon as an enhancement module in
future reconstruction research and products when our API becomes available.

\clearpage
\section{Conclusion}
\label{sec:conclusion}

We presented Plenoptic Condensation (PCon), Quidient's approach to Generalized Scene
Reconstruction (GSR). We showed how unified scene models (Reality Models, aka Relms)
produced by PCon deliver photorealistic appearance, highly accurate geometry, and
relightable meshes without the need for subsequent post-processing. We explained how Relms
are refined to high fidelity by condensing scene elements from lower to higher
representational power.

We captured a highly reflective body panel of a real automobile (a red Fiat 500) indoors
with an iPhone camera. A professional structured light scan was performed to yield 3D
ground truth, including shallow dents in the panel surface. PCon was benchmarked under a
uniform protocol against a SOTA GSR method optimized for appearance accuracy (RT-Splatting)
and a SOTA GSR method optimized for geometric accuracy on specular surfaces (NeRO). PCon
further reconstructed scene elements representing the rest of the scene (other than the
panel) in the near-field and far-field to enable relightability (shown qualitatively with
self-contained consistency checks). The central point is transformational rather than
incremental: because PCon \emph{measures} a unified, disentangled, relightable
representation instead of baking view-dependent appearance, one reconstruction satisfies
GSR needs the field has heretofore satisfied with separate systems or been unable to
satisfy.

\section*{Acknowledgements}
We couldn't produce a document like this without strong support from our customers and our
own team. But we'd particularly like to thank Anav Chaudhary for the hard/smart work
running the Tech Note experiments.

\bibliographystyle{unsrtnat}
\bibliography{references}

\end{document}